\documentclass[sigconf]{acmart}
\AtBeginDocument{%
  \providecommand\BibTeX{{%
    \normalfont B\kern-0.5em{\scshape i\kern-0.25em b}\kern-0.8em\TeX}}}

\usepackage[textsize=tiny]{todonotes}
\usepackage{xcolor}
\usepackage{threeparttable}

\usepackage{amsmath}

\usepackage{amssymb}

\usepackage{mathtools}
\usepackage{amsthm}

\usepackage{enumitem}

\usepackage{hyperref}
\usepackage{makecell}
\usepackage{multirow}

\usepackage{microtype}
\usepackage{graphicx}
\usepackage{subfigure}
\usepackage{booktabs} % for professional tables
\usepackage{multicol}
\usepackage{algorithm}
\usepackage{algpseudocode}
\usepackage{appendix}

\usepackage{mathtools}

\usepackage{amssymb}
\usepackage{amsthm}
\usepackage{color}

\newcommand{\cut}[1]{{}}
\newcommand{\tA}{{\tilde{\vA}}}

\newcommand{\tL}{{\tilde{\vL}}}

\newcommand{\tin}{{\text{in}}}

\newcommand{\vA}{{\mathbf{A}}}

\newcommand{\vD}{{\mathbf{D}}}

\newcommand{\vG}{{\mathbf{G}}}

\newcommand{\vI}{{\mathbf{I}}}

\newcommand{\vL}{{\mathbf{L}}}
\newcommand{\vM}{{\mathbf{M}}}

\newcommand{\vT}{{\mathbf{T}}}

\newcommand{\vX}{{\mathbf{X}}}

\newcommand{\cE}{{\mathcal{E}}}

\newcommand{\cG}{{\mathcal{G}}}

\newcommand{\cL}{{\mathcal{L}}}

\newcommand{\cO}{{\mathcal{O}}}

\newcommand{\cV}{{\mathcal{V}}}

\newcommand{\RR}{\mathbb{R}}

\makeatletter
\let\@@span\span
\def\sp@n{\@@span\omit\advance\@multicnt\m@ne}
\makeatother

\newcommand{\bc}{\begin{center}}
\newcommand{\ec}{\end{center}}

\newcommand{\bdm}{\begin{displaymath}}
\newcommand{\edm}{\end{displaymath}}

\newcommand{\beq}{\begin{equation}}
\newcommand{\eeq}{\end{equation}}

\newcommand{\bfl}{\begin{flushleft}}
\newcommand{\efl}{\end{flushleft}}

\newcommand{\bt}{\begin{tabbing}}
\newcommand{\et}{\end{tabbing}}

\newcommand{\beqn}{\begin{align}}
\newcommand{\eeqn}{\end{align}}

\newcommand{\beqs}{\begin{align*}} % no equation numbers
\newcommand{\eeqs}{\end{align*}}  % no equation numbers

\begin{document}

\title{Efficient End-to-end Language Model Fine-tuning on Graphs}

\author{Rui Xue}
\email{rxue@ncsu.edu}
\affiliation{%
  \institution{North Carolina State University}
  \city{Raleigh}
  \country{USA}
}

\author{Xipeng Shen}
\email{xshen5@ncsu.edu}
\affiliation{%
  \institution{North Carolina State University}
  \city{Raleigh}
  \country{USA}
}

\author{Ruozhou Yu}
\email{ryu5@ncsu.edu}
\affiliation{%
  \institution{North Carolina State University}
  \city{Raleigh}
  \country{USA}
}

\author{Xiaorui Liu}
\email{xliu96@ncsu.edu}
\affiliation{%
  \institution{North Carolina State University}
  \city{Raleigh}
  \country{USA}
}

\renewcommand{\shortauthors}{Rui Xue, et al.}

\begin{abstract}

Learning from Text-Attributed Graphs (TAGs) has attracted significant attention due to its wide range of real-world applications. The rapid evolution of language models (LMs) has 
revolutionized the way we process textual data, which indicates a strong potential to replace shallow text embedding generally used in Graph Neural Networks (GNNs).
However, we find that existing LM approaches that exploit text information in graphs suffer from inferior computation and data efficiency.
In this study, we introduce LEADING, a novel and efficient approach for end-to-end fine-tuning of language models on TAGs. To enhance data efficiency, LEADING efficiently transfers rich knowledge from LMs to downstream graph learning tasks with limited labeled data by employing end-to-end training of LMs and GNNs in a semi-supervised learning setting. To address associated computation efficiency issues, it introduces two techniques: neighbor decoupling targeting LMs and implicit graph modeling targeting GNNs, respectively. Our proposed approach demonstrates superior performance, achieving \textit{state-of-the-art (SOTA)} results on the ogbn-arxiv leaderboard, while maintaining computation cost and memory overhead comparable to graph-less fine-tuning of LMs. Through comprehensive experiments, we showcase its superior computation and data efficiency, presenting a promising solution for various LMs and graph learning tasks on TAGs.

\end{abstract}

\keywords{Graph Neural Networks, Language Models, End-to-end Training}

\begin{abstract}

Learning from Text-Attributed Graphs (TAGs) has attracted significant attention due to its wide range of real-world applications. The rapid evolution of language models (LMs) has 
revolutionized the way we process textual data, which indicates a strong potential to replace shallow text embedding generally used in Graph Neural Networks (GNNs).
However, we find that existing LM approaches that exploit text information in graphs suffer from inferior computation and data efficiency.
In this study, we introduce LEADING, a novel and efficient approach for end-to-end fine-tuning of language models on TAGs. To enhance data efficiency, LEADING efficiently transfers rich knowledge from LMs to downstream graph learning tasks with limited labeled data by employing end-to-end training of LMs and GNNs in a semi-supervised learning setting. To address associated computation efficiency issues, it introduces two techniques: neighbor decoupling targeting LMs and implicit graph modeling targeting GNNs, respectively. Our proposed approach demonstrates superior performance, achieving \textit{state-of-the-art (SOTA)} results on the ogbn-arxiv leaderboard, while maintaining computation cost and memory overhead comparable to graph-less fine-tuning of LMs. Through comprehensive experiments, we showcase its superior computation and data efficiency, presenting a promising solution for various LMs and graph learning tasks on TAGs.

\end{abstract}

\maketitle

\section{Introduction}
\label{sec:intro}

Graph neural networks (GNNs) have been widely used for representation learning on graph-structured data~\citep{hamilton2020graph, ma2021deep}, and they achieve promising state-of-the-art performance on various graph learning tasks, such as node classification, link prediction, and graph classification. Numerous graphs in real-world applications exhibit nodes that are associated with textual attributes, leading to the prevalence of text-attributed graphs (TAGs). TAGs provide a graph-based framework for representing textual data and the connections between them through edges. The fusion of textual attributes and graph topology constitutes valuable information, 
bolstering representation learning in real-world applications such as recommender systems~\citep{jin2023amazon}, citation graphs~\citep{hu2020open, yang2016revisiting}, social networks ~\citep{hamilton2017inductive}, and knowledge graphs~\citep{wang2021wikigraphs}. 

In the context of GNNs, shallow text embeddings such as Bag-of-Words~\citep{harris54} and Word2Vec~\citep{mikolov2013efficient} are usually extracted from raw textual data and used as the numerical node attributes in GNNs due to their superior simplicity and efficiency. However, as they do not fully capture the complex textual semantic features, these approaches inherently restrict the performance of downstream tasks. 
On the other hand, the recursive feature aggregation in GNNs results in the well-known neighborhood explosion problem~\citep{hamilton2017inductive} such that the computation of each node involves its $L$-hop neighbors with $L$ feature aggregation layers.
This not only leads to significant scalability challenges but also limits the exploration of more complex and powerful deep learning techniques such as LMs for the representation learning on TAGs.

Recently, researchers have begun to explore the potential of pre-trained language models (LMs), such as BERT~\citep{devlin2018bert}, DeBERTa~\citep{he2020deberta} and DistilBERT~\citep{sanh2019distilbert}, for representation learning on TAGs due to their unprecedented capabilities in language understanding and generation across a wide range of tasks. 
%%%
The commonly adopted approach follows a \textit{cascaded structure} ~\citep{chen2023exploring}. This entails an initial LM fine-tuning step on downstream tasks such as node classification. 
Subsequently, the text embeddings extracted from the fine-tuned LMs are leveraged as the initial node features for downstream GNNs. 
Although the cascaded structure is efficient, graph structural information is not incorporated in the fine-tuning of LMs, resulting in sub-optimal performance. To address this issue, the \textit{iterative structure} has also been explored for the joint training of LMs and GNNs. 
For instance, GLEM ~\citep{zhao2022learning}
trains LMs and GNNs separately in an iterative manner by generating pseudo labels for each other.
In addition, \textit{self-supervised learning} has also been proposed to enhance LMs fine-tuning by link prediction tasks, exemplified by GIANT~\citep{chien2021node}.

The aforementioned works demonstrate the potential of exploiting LMs on TAGs. However, these approaches still face limitations in \textit{data efficiency} or 
\textit{computation efficiency}. 
First, both cascaded and iterative structures encounter significant data inefficiency. When the labeled data is scarce, these methods struggle to effectively transfer the required knowledge for downstream tasks as the fine-tuning strategies do not utilize labeled data efficiently. 
Second, both iterative structures and the self-supervised learning approach introduce a substantial increase in computational overhead.
This elevated computational cost poses significant scalability challenges, especially when dealing with large-scale datasets.
These shortcomings tremendously limit their applications in transferring the rich knowledge of LMs to facilitate representation learning on TAGs.

In this paper, our aim is to develop an efficient algorithm for fine-tuning LMs that not only effectively adapts LMs to downstream tasks with limited labeled data (\textit{data efficiency}) but also demonstrates superior scalability (\textit{computation efficiency}). We argue that end-to-end training of LMs and GNNs is crucial for achieving data efficiency, as it enables superior knowledge fusion between the two, leveraging the unique advantages of GNN message passing techniques. However, it faces challenges due to scalability (computation efficiency) issues, which we attribute to the giant size of LMs used and the neighbor explosion issue in GNNs. To tackle this, we identify \textit{computation redundancy} as the bottleneck hindering end-to-end training. We further decompose this issue into \textit{encoding redundancy} in LMs and \textit{propagation redundancy} in GNNs. To address these problems, we propose \textit{neighbor decoupling} and \textit{implicit graph modeling} as solutions to alleviate these two redundancy issues respectively. Finally, with the aid of the proposed techniques, we make end-to-end training of LMs and GNNs feasible. Our algorithm demonstrates superior performance, achieving state-of-the-art results on ogbn-arxiv, and exhibits strong scalability comparable to graph-less LM fine-tuning. Therefore, it offers a promising solution for a wide range of LMs and graph learning tasks on TAGs.

\section{Related Work}
\label{sec:relate}

In this section, we will mainly summarize related works exploring language models for learning on TAGs.

\textbf{Basic structure of LMs integrated with GNNs.}  
Several approaches have recently emerged to enhance transformer structures or graph representation techniques. Some of these methods incorporate graph structure information into attention computation~\citep{park2022grpe}, while others introduce orthogonal vectors for node and edge tokens to capture structural nuances~\citep{kim2022pure}. While these enhancements can be effective, they often involve complex attention mechanisms, rendering the direct representation of graph structure a challenging endeavor and significantly increasing the computation complexity of model training.

\textbf{Advanced structure of LMs integrated with GNNs.} 
To address the aforementioned challenges, researchers have explored approaches that combine Language Models (LMs) with graph-based techniques. Notable examples include GLEM~\citep{zhao2022learning}, which employs iterative training as mentioned earlier, and Graphformers~\citep{yang2021graphformers}, which injects GNN layers into LM layers for link prediction. However, these models have their drawbacks. They either rely on a powerful model to generate high-quality soft labels, which necessitate abundant training data, or introduce significant computational overhead. Additionally, there are other approaches like GIANT~\citep{chien2021node}, which uses neighbor prediction to fuse graph into LMs, and E2EG~\citep{dinh2022e2eg}, which incorporates node classification into the joint training process of GIANT. However, these models also face scalability challenges. For example, GIANT utilizes curriculum learning for link prediction, fine-tuning transformers $d$ times, where $d$
represents the depth of the Hierarchical Label Tree (HLT) generated by nodes' TF-IDF text features. 
It entails a significant increase in computation overhead compared to other training strategies \cite{chen2023exploring}. It's important to highlight that, owing to limitations in computational resources, all baseline models choose to fine-tune language models (LMs) rather than large language models (LLMs). As a result, our study primarily revolves around LM fine-tuning, but it can seamlessly integrate with LLMs.

\textbf{Large-scale GNNs.} 
A substantial body of existing research is dedicated to enhancing the efficiency and scalability of large-scale GNNs through innovative designs. These designs encompass sampling methods, pre-computing, and post-computing techniques. Sampling methods employ mini-batch training strategies to reduce computation and memory demands by selectively sampling nodes and edges. They mitigate the neighbor explosion issue through practices such as neighbor sampling~\citep{hamilton2017inductive,chen2018fastgcn,zeng2019graphsaint} or feature memory updating~\citep{fey2021gnnautoscale, xue2023lazygnn}. 
Pre-computing and post-computing methods separate the feature aggregation and prediction models into distinct stages. Pre-computing involves feature aggregation before training~\citep{wu2019simplifying, frasca2020sign, sun2021scalable}, while post-computing includes label propagation after training~\citep{huang2020combining}. 
However, these methods have not been shown to be feasible for the end-to-end training or fine-tuning of LMs.

\section{Methodology}
\label{sec:method}

\renewcommand{\algorithmicrequire}{\textbf{Input:}}
\renewcommand{\algorithmicensure}{\textbf{Output:}}

GNNs have been proven to be data-efficient due to their excellent prediction performance on semi-supervised learning where only very limited labeled data is available. 
%%%
The data efficiency of GNNs can be largely attributed to their ability to integrate node attributes and graph structure information in a unified message-passing framework.
Through end-to-end training, it leverages the scarce labeled data to provide informative supervision for the vast pool of unlabeled nodes. 
However, GNNs' data efficiency comes with the sacrifice of computation efficiency~\citep{hamilton2017inductive}. Hence, most existing approaches exploiting LMs for learning on TAGs fall short in data efficiency and thus fail to effectively adapt the rich knowledge in LMs to downstream graph learning tasks as discussed in Section~\ref{sec:intro} and Section~\ref{sec:relate}. 
We hypothesize that their data inefficiency originates from the fact that existing methods can not fine-tune LMs with graph learning in an end-to-end manner due to the scalability challenges in both LMs and GNNs. 

Building upon the analyses presented above, conducting end-to-end training emerges as a critical factor for enhancing the transfer of knowledge from LMs to the specific downstream tasks. However, the primary challenge we must address is the accompanying scalability (computation efficiency) issue that hinders the application of end-to-end training. We identify the primary challenge as computation redundancy. Consequently, we propose a novel end-to-end fine-tuning strategy (LEADING) to alleviate these redundancies, leading to a highly efficient and scalable solution. Before that, we first introduce the notations as follows.

\textbf{Notations.} 
A graph is represented by $\cG = (\cV, \cE)$ where $\cV = \{v_1, \dots, v_N\}$ is the set of nodes and $\mathcal{E} = \{e_1, \dots, e_M\}$ is the set of edges. 
For a text-attributed graph, each node $v_i$ is associated with a sequential of raw text feature. 
We denote the $d$-dimensional hidden feature vectors of nodes as $\vX \in \RR^{N\times d}$.
The graph structure of $\cG$ can be represented by an adjacency matrix $\vA \in \RR^{N\times N}$, where $\vA_{ij}>0$ when there exists an edge between node $v_i$ and $v_j$, and $\vA_{i,j}=0$ otherwise. 
The symmetrically normalized graph Laplacian matrix is defined as $\tL=\vI-\tA$ with $\tA=\vD^{-1/2}\vA\vD^{-1/2}$ where $\vD$ is the degree matrix.

\subsection{Computation Redundancy in LM-GNN}

\label{sec:redundancy}

Various sampling approaches have been proposed to enhance the scalability and efficiency of GNN training. However, integrating LMs with GNNs in an end-to-end training paradigm introduces its own unique hurdles. This is primarily due to the substantial computational and memory costs associated with LMs, given their large sizes. Additionally, the employment of deep GNNs to capture long-range dependencies poses neighbor explosion challenge to end-to-end LM+GNN training. To tackle the challenges present in both LMs and GNNs, we begin by offering a novel and insightful analysis of computation redundancy within the end-to-end training framework. We identify this redundancy as a crucial factor in achieving successful end-to-end training. We classify the redundancy into encoding redundancy in LMs and propagation redundancy in GNNs. Subsequently, we propose two corresponding techniques, namely neighbor decoupling and implicit graph modeling, to address these issues.

\begin{figure}[htp]
    \centering
    \includegraphics[width=1\linewidth]{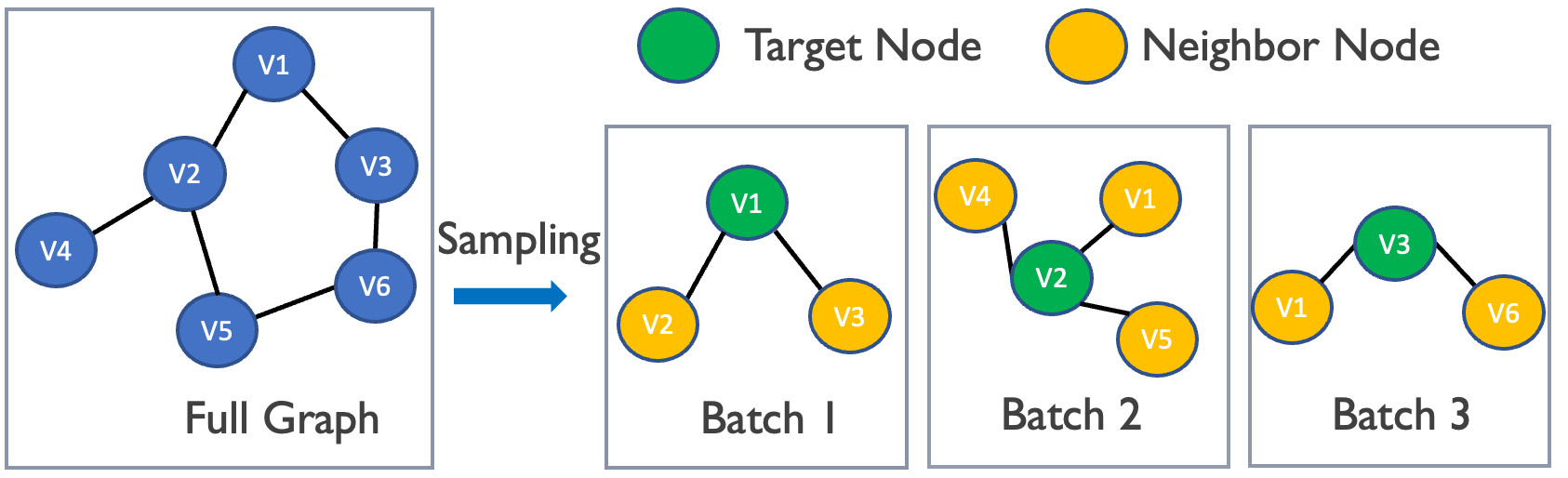}
    \caption{Encoding Redundancy in Mini-batch GNNs.
    }
    \label{fig:encoding}
\end{figure}

\textbf{Encoding Redundancy in LMs.}
In the integration of LMs with GNNs, we have to adopt mini-batch sampling to reduce the computation and memory costs. However, existing sampling strategies of GNNs exhibit heavy redundancy that requires frequently repeated LM encoding of node features, which becomes especially significant due to the immense size of these LMs.
Taking the mini-batch sampling in Figure~\ref{fig:encoding} as an example, the node features need to be encoded by LMs multiple times through every epoch, either as target nodes themselves or as neighbors of other target nodes. For example, $V_1$ serves as a target node in Batch 1 and serves as a neighbor node in Batch 2 and Batch 3. 
However, the LM embedding of the node features will not have notable changes between the mini-batch iterations due to the nature of model fine-tuning.

The above analysis implies that a significant amount of computation on LM encoding is redundant. This redundancy becomes particularly considerable when we employ smaller batch sizes, as typically used in LMs, as well as when we introduce more aggregation layers to capture long-distance information in GNNs. 
According to our statistical analysis on ogbn-arxiv dataset, during each epoch in the training of a 2-layer GNN with neighbor sampling, the node feature of each node is encoded as a target node only once but as a neighbor node 19 times on average when the batch size is 1024 (25 times when the batch size is 64). 
For a 5-layer GNN that requires sampling from 5-hop neighbors, the node feature of each node is encoded 96 times as a neighbor node on average. 
This statistical analysis verifies the LM encoding redundancy in mini-batch GNNs.

\textbf{Propagation Redundancy in GNNs.}
Besides the LM encoding redundancy, there also exists propagation redundancy in GNNs.
As discovered by a recent work~\citep{xue2023lazygnn}, the node embedding in the GNN layers will not change notably over the training iterations but the node information is propagated multiple times repeatedly in each iteration to capture long-distance dependency on graphs. This propagation redundancy causes huge sampling, memory, and computation costs that increase significantly with the number of aggregation layers employed.
Next, we will propose a \underline{L}\underline{a}nguag\underline{e} mo\underline{d}els f\underline{i}ne-tuning o\underline{n} \underline{G}raphs (LEADING) algorithm that tackles the encoding redundancy in Section~\ref{sec:decouple} and propagation redundancy in Section~\ref{sec:implicit}.

\subsection{LEADING in LMs: Neighbor Decoupling}
\label{sec:decouple}

Due to the huge computation and memory cost of LMs, it is imperative to reduce the redundant LM encoding computation.
Our first observation is that for a sampled subgraph, only the target nodes obtain accurate aggregated features and gradients. On the contrary, the major role of neighbor nodes is to facilitate predictions for target nodes but they may not obtain accurate aggregation features and gradients due to their missing neighbors. In other words, the mini-batch neighbor sampling tries to maximally maintain the neighbors of target nodes but the neighbors of neighbors might be out of the batch. As a result, it is feasible to only use the gradient of target nodes to update the LMs.
%%%
The second key observation is that the LM embedding does not change rapidly during the fine-tuning stage, indicating that it is unnecessary to update the LM embedding of neighboring nodes in real-time, considering the significant computational resources that the LM requires.

\begin{figure}[h]
    \centering
    \includegraphics[width=1\linewidth]{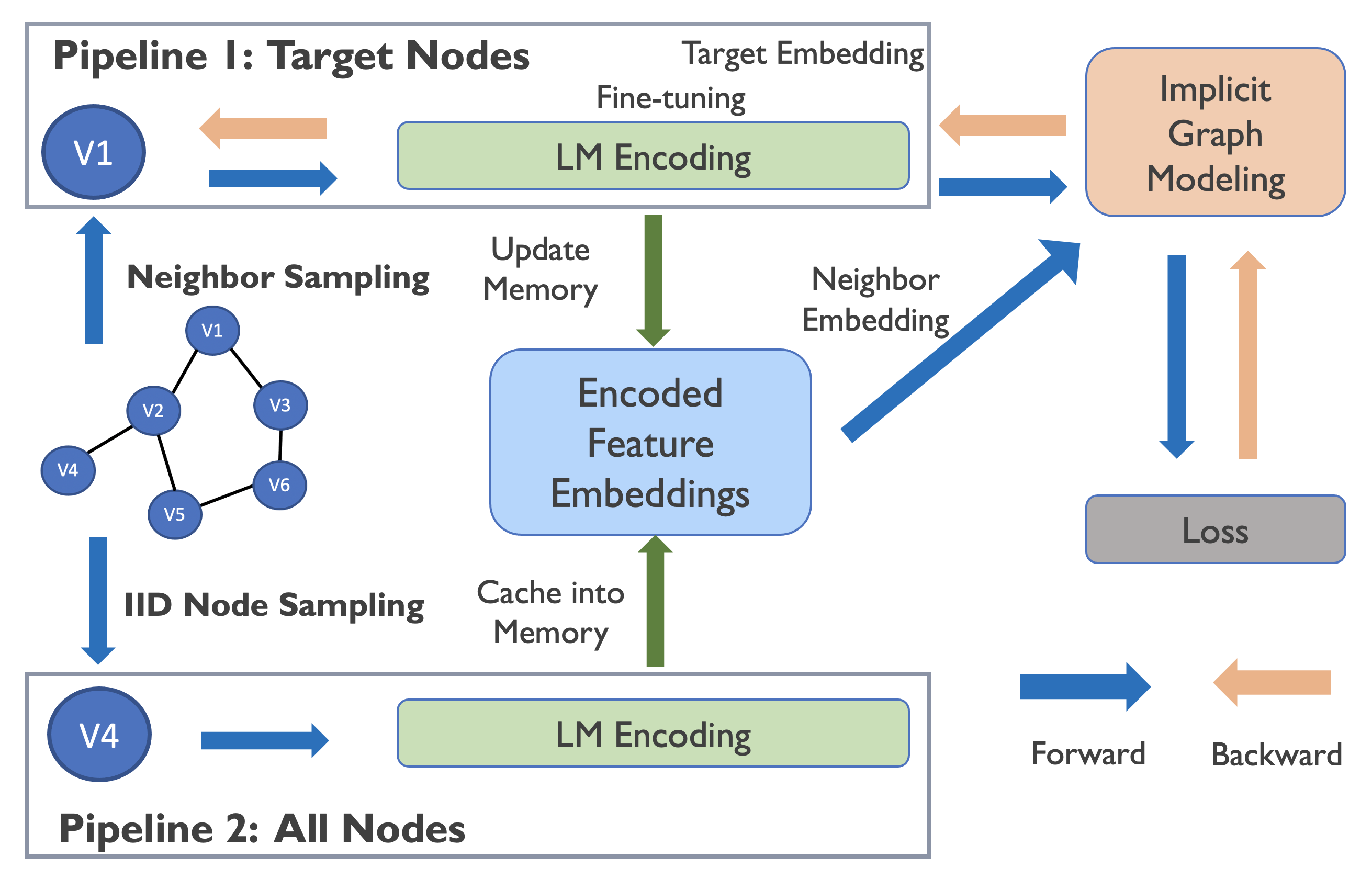}
    \caption{LEADING: two-pipeline training process. 
    (a) a randomly sampled batch is encoded in pipeline 2 and is stored in memory. (b) only the target nodes within the neighbor-sampled batch are encoded with gradients in pipeline 1. (c) Neighbor nodes' embeddings are retrieved and the resulting subgraph is then fed into GNNs. (d) Gradients from target nodes are employed to fine-tune the language models. }
    \label{fig:alg1}
\end{figure}

\textbf{Neighbor Decoupling.}
These key observations motivate us to design a novel training algorithm that fully decouples the LM computation of target nodes and their neighbor node as shown in Figure~\ref{fig:alg1} and Algorithm~\ref{alg}. To reduce the encoding redundancy, we opt to segregate the encoding of target and neighbor nodes into two distinct pipelines. First, for pipeline 2, the LM randomly samples a mini-batch of node text features $\vT_2$ from the whole graph and computes their LM embedding $\vX_2$ without requiring gradient (line 5). It then caches into memory (line 6), which helps facilitate the rapid filling and refreshing of the cache for use in the first pipeline. Clearly, the memory will at least be filled after the first epoch. Following this, neighbor sampling takes place in pipeline 1, where the LM computes the encoding solely for the target nodes $\vT_1$ (line 7) within the current mini-batch. Since only the target nodes require gradients, it can significantly reduce memory costs.
Similarly, the computed LM embedding $\vX_1$ from pipeline 1 (line 8) will be cached in the memory to update the embedding and reduce staleness. Next, the encoded embeddings of neighboring nodes $\vX_{\text{neighbor}}$ are retrieved from the memory bank according to their indexes (line 9) and concatenated with $\vX_1$ to form the node embedding for the entire subgraph (line 10) before being fed into GNNs (line 11). Finally, after computing the loss, backward propagation occurs exclusively within the first pipeline to fine-tune the LMs, as gradients are only required for target nodes.

\begin{algorithm}[h]
    \caption{LEADING Algorithm}    
    \label{alg}       
    \begin{algorithmic}[1] 
    \State \textbf{Input:}   Input Graph $\cG = (\cV, \cE)$, Pre-trained LM $f(\vT, \Theta^0)$
    State \textbf{Output:}  Fine-tuned LM  $f(\vT, \Theta^*)$ 
    \State \textbf{Begin}
    \For{each mini-batch text $\vT_{1}$ in pipeline 1;\\
         each mini-batch text $\vT_{2}$ in pipeline 2}
    \State $\vX_{2} = f(\vT_{2}, \Theta^k)$: Neighbor Nodes Encoding (without gradients)
    \State Cache into Memory $\vM \leftarrow \vX_{2}$
    \State $\vX_{1} = f(\vT_{1}[\text{target}], \Theta^k)$: Target Nodes Encoding (with gradients)
    \State Cache into Memory $\vM \leftarrow \vX_{1}$
    \State Retrieve Neighbor Embeddings $\vX_{\text{neighbor}} \leftarrow \vM$
    \State $\vX_{\text{in}} = \text{Concat}(\vX_{1}, \vX_{\text{neighbor}})$ %: Concatenate corresponding neighbors to the target nodes
    \State $\vX_{\text{out}} = \text{GNN}(\vX_{\text{in}})$
    \State Compute Loss and Gradient Update %Back Propagation
    \EndFor
   \end{algorithmic} 
\end{algorithm} 

This two-pipeline neighbor decoupling approach offers two significant benefits. First, it completely resolves the encoding redundancy problem, as the encoding times of each node in each epoch are controlled by the batch size used in the second pipeline. For instance, if two pipelines take the same batch size, each node feature only needs to be encoded by LM twice, which significantly reduces the computation cost.
Second and more importantly, the memory cost will be significantly reduced since the first pipeline that involves back-propagation training only needs to process target nodes, which are significantly fewer than neighbor nodes. Consequently, the majority of nodes within the mini-batch—the neighbor nodes, which lead to neighbor explosion—need not be a concern. This is the key to achieve LM+GNN end-to-end fine-tuning. 

\subsection{LEADING in GNNs: Implicit Graph Modeling} 
\label{sec:implicit}

While the proposed neighbor decoupling approach significantly reduces memory costs, it is important to note that deep GNN models are typically employed to capture long-distance dependencies in graphs. This introduces a substantial memory overhead, as each intermediate feature embeddings have to be stored to facilitate gradient computation.
Motivated by recent advances in implicit models such as Neural ODE~\citep{chen2018neural}, IGNN~\citep{gu2020implicit}, DEQ~\citep{bai2019deep} as well as the unified view of graph signal denoising~\citep{ma2021unified}, forward propagation can be efficiently computed through a fixed point solver:
\begin{align}
\vX_{*} = \text{Solver}(f_{\theta}(\vX_{*}, \vX_{in})), \lim_{l\rightarrow\infty} \vX_{l} = \vX_{*}
\end{align}
where $f_{\theta}(\vX_{*}, \vX_{in}) = \vX_{*}$ is the implicit function, $\vX_{*}$ represents the fixed point of node embeddings, which corresponds to the final values of the hidden layer in a network when the number of layers tends to infinity $l\rightarrow\infty$. This implicit modeling offers two main advantages: First, the fixed point $\vX_{*}$ represents the equilibrium state achieved after an infinite number of propagations, with only the fixed point and input embeddings needing to be stored for the backward pass, thereby avoiding the storage of any intermediate values and efficiently capturing long-range information. Second, it provides flexibility for the fixed point solver, as the training and inference of implicit deep learning models are independent of the computational trajectory. In other words, any solver can be used to construct the propagation layers. In this work, we have chosen to incorporate the APPNP~\citep{klicpera2018predict} for its role as an iterative solver for fixed points, as it has been proven to effectively alleviate oversmoothing in deep GNN models. The fixed-point embeddings precisely match the exact embeddings as outlined below:

\begin{align}
\vX_{*} &= \alpha \left( \vI - (1-\alpha) \tA \right)^{-1} \vX_{in}, \label{eq:fixed_point}\\
\vX_{l+1} &= (1-\alpha)\tilde{\vA}\vX_{l} + \alpha\vX_{in}, \lim_{l\rightarrow\infty} \vX_{l} = \vX_{*} \label{eq:appnp}
\end{align}

Furthermore, to address propagation redundancy issue, we use $\vX_L^{k-1}$ from the previous training iteration $k-1$ as the initial embedding for the current iteration $k$ since features have been propagated over the graph many times in previous iterations and node embeddings in the GNN layers do not change significantly across iterations, as discussed in Section \ref{sec:redundancy}. In other words, as we know the choice of the initial point affects the convergence, $\vX_L^{k-1}$ serves as a more favorable initialization for the fixed point solver due to its proximity to the fixed point and stability across training iterations. By doing this, only a very few feature aggregation layers (i.e. iteration steps) are required to approximate the fixed-point solution $\vX_*$ in Eq. (\ref{eq:fixed_point}). Then in Algorithm \ref{alg}, the forward propagation can be formulated as:
\begin{align}
\vX_\tin^k &= \text{Concat}\{\vX_1, \vX_{\text{neighbor}}\}, ~~~~ \vX_0^k = \vX_L^{k-1}, \\ 
\vX_{l+1}^k &= (1-\alpha) \tA \vX_{l}^k+\alpha \vX_\tin^k, ~\forall l=0,\dots,L-1,
\end{align}

\noindent where $l$ and $k$ denote the index of layers and training iterations, respectively. The backward propagation can be directly calculated based on the fix-point Eq.(\ref{eq:fixed_point}) by taking derivatives and approximating the matrix inversion iteratively:
\begin{align} 
% \vG_L^k = \frac{\partial \cL}{\partial \vX_L^k}, ~~~
\vG_L^k &= \vG_0^{k-1}, \\
\vG_{l}^k &= (1-\alpha) \tA \vG_{l+1}^k + \alpha \frac{\partial \cL}{\partial \vX_L^k},
~\forall l=L-1, \dots, 0, \label{eq:backward}
\end{align}
\noindent where $G_{L} = \frac{\partial \cL}{\partial \vX_L} ~~ \approx \frac{\partial \cL}{\partial \vX_*} $. Please refer Appendix \ref{proof} for detailed proof. Similarly, the backward propagation starts from the gradient in previous iterations $\vG_0^{k-1}$, which serves as a better initialization. Finally, the gradient of target nodes can be retrieved from $\vG_0^k$ and used for further back-propagation in the LM $f(\vT_1, \Theta^k)$.
After the end-to-end fine-tuning, we can utilize the fine-tuned LM to generate feature embedding, which serves as the initial embedding for any downstream GNNs and graph learning tasks.

\subsection{Computation Complexity Analysis}
\label{sec:complexity}

\textbf{LM Complexity Analysis.} 
Suppose $N$ is the total number of nodes and $C$ is the computation complexity of encoding one node feature by LMs. Additionally, let $P$ stand for the average encoding times per node in LEADING and $Q$ is the average encoding time per node in existing scalable GNNs, such as GraphSAGE or GAS \cite{fey2021gnnautoscale}. Then, the total computation complexity of LM encoding is $\cO(PNC)$ for LEADING and $\cO(QNC)$ for other baselines. Given that $P \ll Q$ (where $P$ is constant and equals 2 in our experiments), as discussed in Section \ref{sec:redundancy} and shown in Figure \ref{fig:encode_gnn}, LEADING evidently achieves better computation efficiency.

Regarding memory complexity, suppose $\cO(S)$ is the memory complexity for executing forward and backward propagation per node. 
For the mini-batch sampling, suppose $T$ and $B$ are the batch sizes of target nodes and neighbor nodes, respectively. Typically, we have $B \gg T$. Then the total memory complexity for LEADING is $\cO(TS)$, which is the same as training LMs without using graphs. It is much lower than the normal GNN training strategy whose memory complexity is $\cO((B+T)S)$. These complexity analyses indicate the intriguing scalability of LEADING in the LM phase.

\textbf{GNN Complexity Analysis.}
Due to the variety of downstream GNN structures used in baselines, which are not the focus of this section, we use the GraphSAGE an an example to simplify the complexity analysis and directly demonstrate the superior efficiency of our model. Suppose $N$ is the total number of nodes, $L$ is the number of propagation layers, and $H$ is the size of hidden units, and $R$ is the number of sampled neighbors of each node. The time complexity of GraphSAGE is $\cO(R^{L}NH^2)$ \cite{li2021training}. It is worth noting that LEADING has fewer layers $\Tilde{L} \ll L$ compared to existing approaches as described in Section~\ref{sec:implicit}. This reduction in the number of layers contributes to lowering the overall computation cost in feature aggregation.

For memory complexity, regular GraphSAGE requires $\cO(R^{L}TH)$ to store the intermediate state at each feature aggregation layer. This complexity grows exponentially with the number of layers. However, our algorithm achieves a memory complexity of $\cO(TH)$ because we utilize implicit gradient modeling, which does not requires the storage of feature in intermediate layers. Therefore, the memory cost is independent of the number of aggregation layers. This indicates a significant reduction in terms of memory cost.

\vspace{-0.15in}
\begin{table}[!htb]
\begin{center}
\renewcommand{\arraystretch}{1.2}
\caption{Complexity Analysis}
\label{complexity}
\vspace{-0.15in}
\resizebox{0.83\linewidth}{!}{%
\begin{threeparttable}
\begin{tabular}{l|c|c}
\toprule
\multirow{1}{*}{\textbf{Method}} & \multicolumn{1}{c|}{\bf Time} & \multicolumn{1}{c}{\bf Memory} \\
\toprule
LM+SAGE & $\mathcal{O}(QNC + R^{L}NH^2)$ & $\mathcal{O}((B+T)S + R^{L}TH)$ \\
Cascaded & $\mathcal{O}(NC + R^{L}NH^2)$ & $\mathcal{O}(TS + R^{L}TH)$ \\
GLEM & $\mathcal{O}(K(NC + R^{L}NH^2))$ & $\mathcal{O}(TS + R^{L}TH)$ \\
LEADING & $\mathcal{O}(NC + R^{\widetilde{L}}NH^2)$ & $\mathcal{O}(TS + TH)$ \\
\bottomrule
\end{tabular}
\begin{tablenotes}
  \item $K$ is the number of iterative training rounds of LM and GNN in GLEM; $LM+SAGE$ denotes training the language model and GNN jointly.
\end{tablenotes}
\end{threeparttable}
}
\vspace{-0.2in}
\end{center}
\end{table}

\section{Experiment}
\label{sec:exp}

In this section, we present experiments to demonstrate the superior data efficiency and computation efficiency of LEADING.
In particular,
we try to answer the following questions: 
(Q1) Data efficiency: can our LEADING algorithm transfer the knowledge from LMs to downstream graph learning tasks effectively with limited training data? (Section~\ref{sec:performance}) and (Q2) Computation efficiency: can our LEADING algorithm be more scalable compared with other fine-tuning paradigms and GNNs? (Section~\ref{sec:efficiency}, \ref{sec:scalability})

\textbf{Datasets.} 
We conduct experiments on both small and large text-attributed graph datasets including Cora~\citep{mccallum2000automating}, PubMed~\citep{sen2008collective}, ogbn-arxiv and ogbn-products~\citep{hu2020open}. 
We evaluate the effectiveness of LM fine-tuning by taking semi-supervised node classification problems as the downstream tasks.
We randomly split the data into training/val/test sets 10 times for Cora and PubMed and report the mean and variance of accuracy following existing works~\citep{kipf2016semi}. 
We adopt the default labeling ratios of these datasets, i.e., 20 training nodes per class for Cora and PubMed (low labeling rate) and 53.7\% for ogbn-arxiv (high labeling rate). Additionally, we present a specific case with a high labeling rate on Cora. Please refer to Appendix \ref{high} for details.

\textbf{Baselines.} 
We compare the proposed LEADING algorithm with some classic baseline and a set of LM fine-tuning strategies:
(1) \textit{Shallow Embedding:} Default shallow embeddings provided by PyG ~\citep{fey2019fast}; (2) \textit{Pre-trained LMs:} LMs function as simple encoders without fine-tuning on labeled data, and the resulting feature embeddings are used as inputs for downstream GNNs;
(3) \textit{Supervised-FT LMs:} LMs are directly fine-tuned using the labeled data under the supervised setting. Subsequently, the text embedding generated by the fine-tuned LMs is used as the node embedding for downstream GNNs;
(4) \textit{GIANT \& GLEM:} We choose GIANT~\citep{chien2021node} and GLEM~\citep{zhao2022learning} as the major baselines since they exhibit the excellent performance among all existing works\cite{chen2023exploring}. Moreover, GLEM is a representative method of iterative training strategy, while GIANT is a representative method of self-supervised training strategy.
It is worth noting that due to the high training costs of GIANT, we only use pre-trained features provided by their official repository. 
(5) \textit{GraD \& TAPE:} We have incorporated another newly proposed model, GraD\cite{mavromatis2023train}, which has demonstrated strong performance on the OGB leaderboard and has publicly available code. Additionally, we include TAPE\cite{he2023explanations}, which employs LLMs as an enhancer to generate explanations to aid in boosting downstream performance. We adhere to their experimental settings, including the datasets used (ogbn-arxiv). It's worth noting that TAPE is entirely orthogonal to our algorithm; hence, we also report the performance of LEADING augmented with TAPE features.
(6) \textit{GraphFormers \& Grenade:} We also include two end-to-end joint training paradigms: classical models like GraphFormers\cite{yang2021graphformers} and state-of-the-art models like Grenade\cite{li2023grenade}. To ensure consistency with the experiment settings proposed in original papers, we conduct comparisons on the ogbn dataset.

\textbf{Evaluation setting.} 
For LMs, in order to ensure a fair comparison with the baselines across different datasets, we include the same LMs as used in their respective studies. Specifically, we employ BERT~\citep{devlin2018bert} as utilized in GIANT, DeBERTa~\citep{he2020deberta} as applied in GLEM, and one lightweight variant, DistilBERT~\citep{sanh2019distilbert}. Additionally, to further validate the effectiveness of our algorithm, we conduct additional experiments involving a larger LM, one of the most popular decoder-only models known as GPT-2. Results and detailed analysis can be found in Appendix \ref{GPT}.

To evaluate the effectiveness of LMs fine-tuning, we extract the CLS (classification) embedding from the last hidden states of fine-tuned encoder-only LMs as the text embeddings, aligning with the configuration used in the baseline models. For downstream GNNs, we conduct performance comparisons on Cora and Pubmed using two classic GNNs, namely GCN~\citep{kipf2016semi} and GAT~\citep{velivckovic2017graph}. In the case of ogbn-arxiv dataset, we employ GCN and Rev-GAT~\citep{li2021training}. For the ogbn-products dataset, we opt for GraphSage~\citep{hamilton2017inductive} and GAMLP~\citep{zhang2022graph} following existing works~\citep{chen2023exploring,zhao2022learning}. We perform all hyperparameter tuning following baselines. We also provide an ablation study on important hyperparameters in the Appendix \ref{hyperparameters}.

\subsection{Prediction Performance}
\label{sec:performance}

\textbf{Major baselines} We evaluate the effectiveness of LM fine-tuning by comparing the prediction accuracy on downstream GNNs. 
From Table~\ref{tab:accuracy-llm}, we can make the following observations:

\begin{table*}[!htb]
\caption{Prediction accuracy (\%) of LM fine-tuning strategies. 
The \textbf{best} are marked as bold.}
\vspace{-0.15in}
\label{tab:accuracy-llm}
\begin{center}
\renewcommand{\arraystretch}{1.2}
\setlength{\arrayrulewidth}{1.5pt}
\resizebox{0.8\textwidth}{!}{%
\begin{threeparttable}
\begin{tabular}{lcccccccc}
\toprule
\textbf{\textsc{Dataset}} & \multicolumn{2}{c}{\bf Cora}  &\multicolumn{2}{c}{\bf Pubmed} &\multicolumn{2}{c}{\bf Arxiv} &\multicolumn{2}{c}{\bf Products}
\\ \toprule
\textbf{\textsc{Method}} & \multicolumn{1}{c}{\bf GCN} & \multicolumn{1}{c}{\bf GAT} &\multicolumn{1}{c}{\bf GCN} & \multicolumn{1}{c}{\bf GAT} &\multicolumn{1}{c}{\bf GCN}& \multicolumn{1}{c}{\bf Rev-GAT} &\multicolumn{1}{c}{\bf SAGE}& \multicolumn{1}{c}{\bf GAMLP}\\
\midrule
% Default Feature
Shallow Embedding  & $82.0 \pm 0.7$ & $82.3 \pm 0.7$ & $78.9 \pm 1.0 $ & $77.7 \pm 0.9 $ & 71.7 & 73.6 & 79.7 & 83.5\\
\midrule
Pre-trained DeBERTa  & $48.5 \pm 1.9$ & $51.0 \pm 1.2$ & $62.0 \pm 0.1$ & $62.6 \pm 0.3$ & 45.7 & 47.8 & 62.0 & 82.4\\
\midrule
% FT BERT Feature      
Supervised-FT BERT  & $77.3 \pm 1.7$ & $78.2 \pm 1.4$ & $68.6 \pm 1.8$ & $68.6 \pm 1.4$ & 73.1 & 73.8 & 81.8 & 79.8\\
Supervised-FT DistilBERT  & $ 79.5\pm1.5 $ & $ 79.2\pm1.7 $ & $ 72.8\pm1.2 $ & $ 72.6\pm1.1 $ & 73.0 & 73.7 & 81.5 & 80.4\\
% FT DeBERTa Feature 
Supervised-FT DeBERTa & $59.2 \pm 1.2$ & $57.4 \pm 2.0$ & $62.1 \pm 0.1$ & $61.6 \pm 0.1$ & 74.7 & 75.8 & 82.2 & 80.7\\
\midrule
GIANT (BERT)   & --- & --- & --- & --- & 73.3 & 75.9 & 83.1 & 83.7\\
GLEM (DeBERTa)\tnote{*}  & $59.2 \pm 1.2$ & $57.4 \pm 2.0$ & $62.1 \pm 0.1$ & $62.6 \pm 0.3$ & 75.9 & 76.9 & 83.2 & 85.1\\
\midrule
LEADING (BERT)   & $80.5 \pm 0.4$ & $81.6 \pm 0.3$ & $79.1 \pm 0.5$ & $79.0 \pm 1.0$ & 73.8 & 74.8 & 83.8 & 85.7\\
LEADING (DistilBERT)   & $\mathbf{82.5 \pm 0.5}$ & $\mathbf{83.0 \pm 0.5}$ & $79.4 \pm 0.4$ & $79.2 \pm 0.8$ & 73.5 &  74.3 & 83.7& 85.3\\
LEADING (DeBERTa)   & $80.6 \pm 0.3$ & $81.4 \pm 0.6$ & $\mathbf{79.5 \pm 0.8}$ & $\mathbf{79.3 \pm 0.6}$ & \textbf{76.1} & \textbf{77.6} & \textbf{84.1} & \textbf{86.5}\\

\bottomrule
\end{tabular}
\begin{tablenotes}
    \item[*]  In our investigation, we found that in the low labeling ratio case (such as Cora and PubMed), GLEM achieves its highest accuracy when the ratio of pseudo labels is set to 0. In this unique case, GLEM essentially operates similarly to Pre-trained/Supervised Fine-tuning Language Models, relying solely on truth labels for training. This observation aligns with \citet{chen2023exploring}.
\end{tablenotes}
\end{threeparttable}
}
\vspace{-0.15in}
\end{center}
\end{table*}

\begin{itemize}
[leftmargin=0.15in] 
\setlength\itemsep{0em}
    \item In the low labeling setting (Cora and Pubmed), LEADING outperforms all other LM fine-tuning strategies. Notably, compared with Supervised-FT DeBERTa, LEADING significantly boosts the performance of the DeBERTa from $59.2\%$ to $80.6\%$ for GCN and from $57.4\%$ to $81.4\%$ for GAT on Cora. A similar improvement can be observed on PubMed as well. 

    \item Comparing Pre-trained DeBERTa with Supervised-FT DeBERTa, fine-tuning without an end-to-end manner does not provide significant benefits in the low-labeling setting (Cora and PubMed) due to the scarcity of labeled data. However, it can prove more beneficial as the volume of training data increases (ogbn-arxiv).

    \item We found that all baselines perform poorly in low labeling rate settings, revealing a gap compared to shallow embeddings. We conclude that when training samples are limited, these methods fail to transfer sufficient pre-trained knowledge from LMs to downstream GNN tasks, which supports our claim in Section \ref{sec:intro} and is consistent with existing works \cite{chen2023exploring}. However, LEADING does not suffer from this limitation. Thanks to our proposed end-to-end fine-tuning, it effectively fills the performance gap observed in baselines and performs the best in all scenarios, highlighting its necessity.

    \item In the high labeling setting (e.g., ogbn-arxiv), LEADING also achieves strong performance. For DeBERTa, LEADING achieves $76.1\%$ and $77.6\%$ accuracy for GCN and Rev-GAT, which are better than GLEM ($75.9\%$ and $76.9\%$), a model that has proven to be very strong in the high labeling setting~\citep{chen2023exploring}. In the case of ogbn-products, LEADING achieves $84.1\%$ for SAGE and $86.5\%$ for GAMLP, outperforming all other baseline methods. We also include statistical significance tests to illustrate the substantial performance improvement in Appendix \ref{significant}.
    
    \item It should be noted that LEADING achieves these remarkable performances with much better computation efficiency and scalability as will be discussed in Section~\ref{sec:efficiency}.

\end{itemize}

\noindent\textbf{SOTA performance} From Table \ref{sota}, our approach has achieved best performance on the OGB leaderboard\footnote{\url{https://ogb.stanford.edu/docs/leader_nodeprop/}} for ogbn-arxiv surpassing all recent baselines without the need for additional LLMs as enhancers or augmentations. Notably, by incorporating augmented features from TAPE\cite{he2023explanations}, we effortlessly achieved the state-of-the-art performance. This underscores LEADING's superior capability in terms of performance and the generality of combining it with other techniques. It validates that end-to-end training can indeed be beneficial, consistent with our analysis in Section \ref{sec:method}. Importantly, it pushes the boundaries of graph machine learning on TAGs, emphasizing the necessity of this work.

\begin{table}[!htb]
\centering
\renewcommand{\arraystretch}{1.2}
\caption{Prediction accuracy (\%) comparison with other baselines on Arxiv}
\label{sota}
\vspace{-0.15in}
\resizebox{0.9\linewidth}{!}{%
\begin{tabular}{l|cccc}
\toprule
\multirow{1}{*}{\textbf{Method}} & \multirow{1}{*}{\textbf{GraD}} & \multirow{1}{*}{\textbf{TAPE}} & \multirow{1}{*}{\textbf{LEADING}} & \multicolumn{1}{c}{\textbf{LEADING + TAPE}} \\
% & & & & \multicolumn{1}{c}{\textbf{+ TAPE}}\\
\midrule
Rev-GAT & 77.2 & 77.5 & 77.6 & \textbf{78.2} \\
\bottomrule
\end{tabular}
}  
\vspace{-0.2in}
\end{table}

\noindent\textbf{Other joint training baselines} To further demonstrate the superior performance of our LEADING algorithm, we compared it against other end-to-end models, including classical models like GraphFormers\cite{yang2021graphformers} and state-of-the-art models like Grenade\cite{li2023grenade}. The results show that LEADING outperforms both baselines across all datasets, further proving its effectiveness in achieving knowledge transfer between GNNs and language models. For detailed results and analyses, please refer to Appendix \ref{e2e}.

\begin{table*}[!htb]
\caption{Scalability comparison between different LM fine-tuning strategies}
\vspace{-0.15in}
\label{tab:efficiency}
\begin{center}
\renewcommand{\arraystretch}{1.2}
\setlength{\arrayrulewidth}{1.5pt}
\resizebox{0.7\textwidth}{!}{%
\begin{tabular}{lcccc}
\toprule
\textbf{\textsc{Dataset}}
& \multicolumn{2}{c}{\bf Arxiv} &\multicolumn{2}{c}{\bf Products}
\\ \toprule
\textbf{\textsc{Method}} & \multicolumn{1}{c}{\bf Memory(GB)} & \multicolumn{1}{c}{\bf Running Time(S)} & \multicolumn{1}{c}{\bf Memory(GB)} & \multicolumn{1}{c}{\bf Running Time(S)}\\
\midrule
Supervised-FT BERT    & 11.5 & 8400  & 14.0 & 19900\\
Supervised-FT DeBERTa       & 13.6 & 12200 & 25.1 & 24640\\
\midrule
GIANT (BERT)    & N/A & N/A & N/A & N/A \\
GLEM (DeBERTa)     & 13.6 & 67634 & 25.2 & 137760\\
\midrule
LEADING (BERT)    & 11.7 & 15241 & 14.9 & 35748\\
LEADING (DeBERTa)    & 13.9 & 22226 & 25.9 & 44388\\
\bottomrule
\end{tabular}
}
\vspace{-0.1in}
\end{center}
\end{table*}

\subsection{Efficiency Analysis}
\label{sec:efficiency}

In this section, we investigate the computation efficiency and scalability during the LM fine-tuning stage. 
We select BERT and DeBERTa as the LM architectures since they are used in the baselines of GIANT and GLEM. The results in Table~\ref{tab:efficiency} reveal the following noteworthy observations:

\begin{itemize}
[leftmargin=0.15in] 
\setlength\itemsep{0em}

\item Notably, the proposed LEADING achieves a memory cost that is nearly identical to Supervised-FT LM training without using graphs, which aligns with our expectations. This is attributed to the proposed neighbor decoupling and implicit graph modeling as introduced in Section~\ref{sec:method}.

\item The iterative training strategy, such as GLEM, and self-supervised training strategy, such as GIANT, exhibit significantly higher memory costs or longer running times compared to the cascaded structure, such as Supervised-FT. Specifically, since all our experiments were conducted using a single GPU, replicating the results of GIANT, which originally used 8 V100 GPUs for training, posed challenges. To replicate their results, we attempted to scale up the batch size by a factor of 8, resulting in out-of-memory (OOM) issues. While reducing the batch size is a possible solution, it significantly prolongs the running time and lowers performance. Therefore, we report N/A in the table. This orders-of-magnitude larger computational overhead can be attributed to the fact that GIANT employs a multi-level fine-tuning approach as discussed in Section \ref{sec:relate}, which results in a significant increase in computation overhead compared to other training strategies \cite{chen2023exploring}.
\item Hyperparameters tuning: LEADING does not require additional hyperparameters tuning, which further strengthens its efficiency and simplicity. However, GLEM is more complicated and requires additional hyperparameters tuning such as the ratio of generated pseudo labels, the number of iterations of the EM-Step, etc.

\item The running time of LEADING is around $0.8$ times higher than that of Supervised-FT, which is reasonable since the two pipelines are run in a sequential manner on the same GPU but it can be easily reduced by parallel computing. The memory cost and running time align well with our 
complexity analysis in Section~\ref{sec:complexity}.

\end{itemize}

\subsection{Scalability Comparison}
\label{sec:scalability}

In this section, we conduct a comparative analysis between our approach and other scalable GNN backbones, considering both performance and efficiency metrics. We choose two widely-used scalable GNNs, namely GraphSAGE~\cite{hamilton2017inductive} and one of the state-of-the-art scalable algorithm 
GNNAutoScale~\cite{fey2021gnnautoscale}, as our primary baselines.
To expedite computations, we employ DistilBERT~\citep{sanh2019distilbert} paired with a 2-layer APPNP in an end-to-end fashion on Cora, Pubmed and ogbn-arxiv. We use same hyperparameters to ensure a fair comparison.

\begin{figure}[h]
  \centering
  \includegraphics[width=0.5\linewidth]{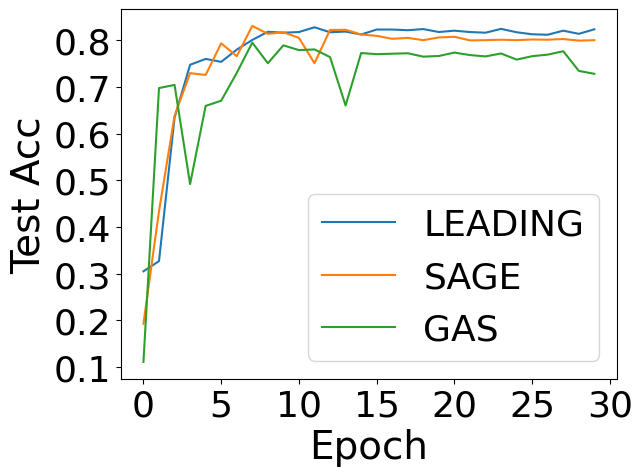}
  \vspace{-0.1in}
  \caption{Convergence comparison} 
  \label{fig:convergence}
  \vspace{-0.2in}
\end{figure}

As shown in Table~\ref{gnns}, LEADING can achieve comparable or superior results compared to the coupling method, validating the rationale of LEADING as discussed in Section~\ref{sec:decouple}. And all three approaches converge at the same speed, as illustrated in Figure~\ref{fig:convergence}. In terms of memory cost, LEADING stands out as significantly more efficient and the only model capable of end-to-end training on ogbn-arxiv. This is because even state-of-the-art scalable GNN training algorithms such as GAS still encode target nodes and their first-hop neighbors together, which encounter out of memory issues. On the contrary, our proposed algorithm achieves a novel breakthrough by completely decoupling target nodes from their neighbors.

\begin{table}[!h]
\begin{center}
    \renewcommand{\arraystretch}{1.2}
    \caption{Performance and corresponding memory cost comparison between normal LM-GNN training and LEADING.}
    \vspace{-0.15in}
    \label{gnns}
    \resizebox{0.85\linewidth}{!}{%
    \begin{tabular}{l|ccc}
    \toprule
    \textbf{Method /Accuracy(\%)} & \multicolumn{1}{c}{\bf Cora} & \multicolumn{1}{c}{\bf Pubmed} & \multicolumn{1}{c}{\bf Arxiv} \\
    \toprule
    LM $+$ GraphSAGE & $82.0 \pm 0.6$ & $80.3 \pm 0.4$ & OOM \\
    LM $+$ GAS & $81.3 \pm 0.3$ & $79.5 \pm 0.2$ & OOM \\
    LEADING & $81.9 \pm 0.7$ & $80.1 \pm 0.4$ & $74.3$ \\
    \midrule
    \textbf{Method /Memory(MB)} & \multicolumn{1}{c}{\bf Cora} & \multicolumn{1}{c}{\bf Pubmed} & \multicolumn{1}{c}{\bf Arxiv} \\
    \toprule
    LM $+$ GraphSAGE & $21021$ & $35070$ & OOM \\
    LM $+$ GAS & $15844$ & $19613$ & OOM \\
    LEADING & $2398$ & $2399$ & $26557$ \\
    \bottomrule
    \end{tabular}
    }
    \vspace{-0.15in}
\end{center}
\end{table}

\begin{figure*}[!t]
  \begin{minipage}[t]{0.55\textwidth}
    \centering
    \hspace{-1.3in}
    \includegraphics[width=0.45\textwidth]{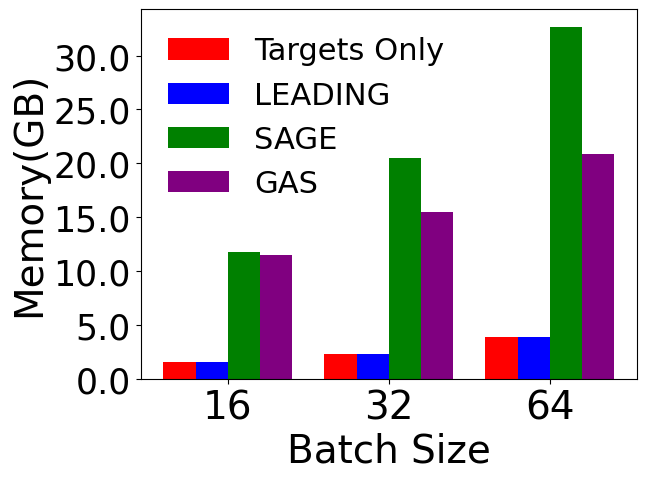}
    \includegraphics[width=0.45\textwidth]{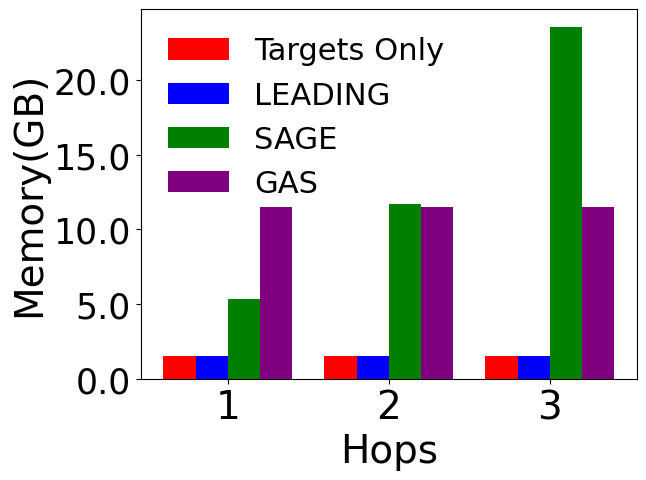}
    \vspace{-0.1in}
    \caption{Memory Cost\hspace{1.5in}}
    \label{fig:memory_gnn}
  \end{minipage}
  \hspace{-1.5in}
  \begin{minipage}[t]{0.55\textwidth}
    \centering
    \includegraphics[width=0.45\textwidth]{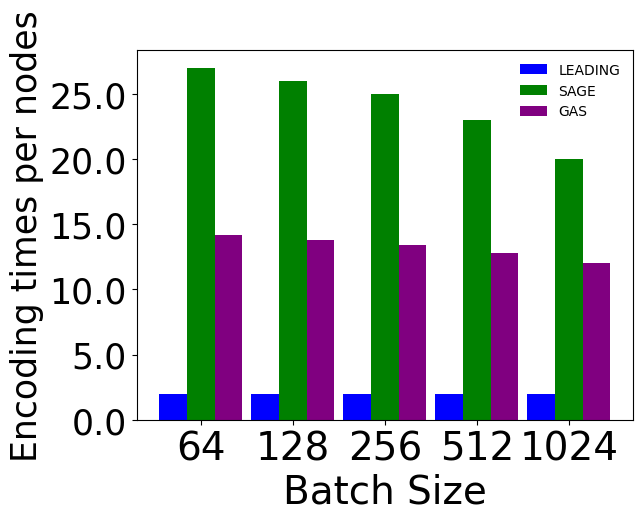}
    \includegraphics[width=0.45\textwidth]{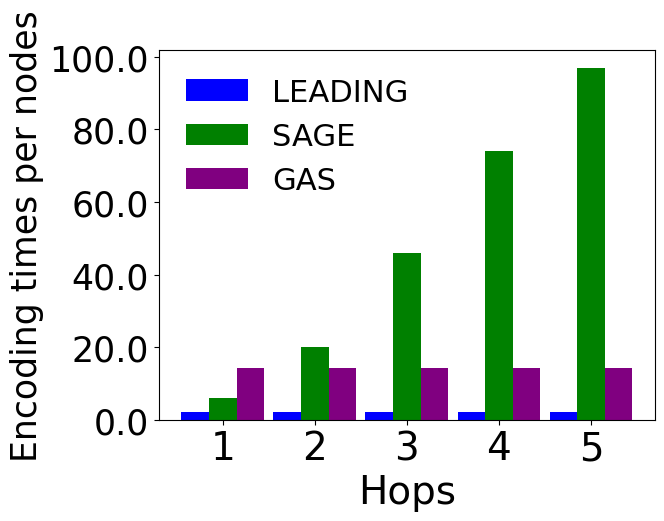}
    \vspace{-0.1in}
    \hspace{-1.2in}
    \caption{Encoding Times}
    \label{fig:encode_gnn}
  \end{minipage}
\end{figure*}

We also present a study on the memory usage on Cora and the average LM encoding times of each node on ogbn-arxiv. We assess the memory usage based on two key factors: (1) different batch sizes while keeping 2 hops neighbors (sampling 10 neighbors for the first hop and 5 for the second hop for each node); (2) varying numbers of hops (sampling 10 neighbors for the first hop and 5 for the following hops for each node) while keeping a fixed batch size. 

The memory usage in Figure \ref{fig:memory_gnn} reveals that a substantial portion of the computation cost is associated with the encoding of neighboring nodes. Despite GAS achieving improved memory efficiency, especially with a large batch size, and maintaining nearly the same cost as GNNs grow deeper, it still demands significantly more memory than LEADING due to its coupling of 1-hop neighbors. Importantly, LEADING sustains a consistent memory cost comparable to Supervised-FT LMs (``Targets Only'') and remains independent of the number of neighbors included, underscoring a notable scalability advantage.

The average LM encoding times of each node in Figure~\ref{fig:encode_gnn} indicate a considerable level of computational redundancy and this redundancy is affected by both batch size and number of neighbors. The results also show that our LEADING algorithm ensures that in each epoch, each node is only encoded by LM twice, significantly minimizes computation redundancy due to the decoupled computation of target nodes and neighbor nodes. It brings significant cost reduction compared to other scalable GNN training strategies.

\subsection{Ablation Study}
\label{sec:ablation}

\textbf{The necessity and importance of two mechanisms.} 
We provide an ablation study to elucidate the individual contributions of the techniques in terms of the computation efficiency. We employ the APPNP as GNN backbones and DeBERTa as the LM for our experiments on ogbn-arxiv. We conduct tests with different layer $L$ in four settings: (1) DeBERTa + APPNP (batch size 32); (2) DeBERTa + APPNP + Neighbor Decoupling (denoted as ND) (batch size 32); (3) DeBERTa + Implicit Modeling (denoted as IM) + Neighbor Decoupling (batch size 32); (4) DeBERTa + Implicit Modeling + Neighbor Decoupling (batch size 128).

For consistency, we use a batch size of 32 or 128 with two-hop neighbors (5 neighbors for each hop). The obtained results in Table \ref{tab:individual} reveal that both neighbor decoupling and implicit modeling play crucial roles in reducing memory costs.  
Regarding neighbor decoupling, it significantly reduces memory costs without sacrificing performance. The additional running time is attributed to the sequential execution of two pipelines in our algorithm. This issue can be easily mitigated through parallel training, as described in Section~\ref{sec:efficiency}. Due to the relatively small target nodes batch size of 32 used and only target nodes requiring gradients, simply adding layers will not lead to an obvious increase in memory usage for rows 5 to 8. Considering the implicit graph modeling, it achieves the same performance as 10 layers APPNP with only 2 propagation layers. Consequently, implicit modeling not only significantly reduces running time but also decreases memory costs. Furthermore, performance can be further enhanced by increasing the batch size. These findings underscore the advantages of implicit modeling and its ability to capture long-distance dependencies in graphs with very few propagation layers. Two mechanisms both play important roles in LEADING.

\begin{table}
\begin{center}
\caption{Individual contribution towards the efficiency.}
\label{tab:individual}
\vspace{-0.2in}
\renewcommand{\arraystretch}{1.5}
\setlength{\arrayrulewidth}{1.5pt}
\resizebox{1.05\linewidth}{!}{%
\begin{tabular}{l|ccc}
\toprule
\textbf{\textsc{Methods}} &\multicolumn{1}{c}{\bf Accuracy (\%)} &\multicolumn{1}{c}{\bf Time/Epoch (s)} &\multicolumn{1}{c}{\bf Memory(MB)} \\
\midrule
% Default Feature
LM+APPNP(L=2, BS:32) & 70.8 & 5770 & 47243\\
LM+APPNP(L=3, BS:32) & N/A & N/A & OOM\\
LM+APPNP(L=5, BS:32) & N/A & N/A & OOM\\
LM+APPNP(L=10, BS:32) & N/A & N/A & OOM\\
\midrule
LM+APPNP + ND(L=2, BS:32) & 70.8 & 12117 & 8522\\
LM+APPNP + ND(L=3, BS:32) & 71.3 & 18215 & 8522\\
LM+APPNP + ND(L=5, BS:32) & 72.1 & 30311 & 8522\\
LM+APPNP + ND(L=10, BS:32) & 72.8 & 60680 & 8522\\
\midrule
LM + ND + IM(L=2, BS:32) & 72.9 & 12117 & 6259\\
LM + ND + IM(L=2, BS:128) & 73.6 & 2996 & 25103\\
\bottomrule
\end{tabular}
}
\vspace{-0.35in}
\end{center}
\end{table}

\textbf{Link Prediction and PEFT.}
To emphasize the effectiveness of our LEADING algorithm, we conducted additional experiments on link prediction tasks. The results demonstrate that LEADING exhibits performance advantages over baselines on all datasets, especially in scenarios with limited training data, reinforcing our conclusion from node classification and highlighting the universality of our algorithm. Additionally, to further illustrate the versatility of LEADING, we demonstrate the successful integration of the LEADING algorithm with existing Parameter Efficient Fine-Tuning (PEFT) approaches, such as LoRA~\citep{hu2021lora}, which further enhances performance. Please refer to Appendices \ref{link} and \ref{peft} for detailed results and analyses.

\section{Conclusion}

Exploring the potential of pre-trained LMs for representation learning on TAGs has been of significant interest in recent years. However, it comes with significant efficiency issues in the integration of powerful LMs and GNNs. 
In this work, we revisit and analyze the limitations of existing approaches with a special focus on data efficiency and computation efficiency. 
To resolve these limitations, this work develops a novel and efficient LM-GNN end-to-end fine-tuning algorithm (LEADING) that not only effectively adapts LMs to downstream graph learning tasks with limited labeled data but also exhibits strong scalability and efficiency. Comprehensive experiments validate its superior prediction performance and efficiency in both low labeling ratio and high labeling ratio settings. The proposed algorithm offers a promising solution for the end-to-end integrating LMs and GNNs in various impactful applications.

\bibliographystyle{ACM-Reference-Format}
\bibliography{ref}

\newpage

\addcontentsline{toc}{section}{Appendix} % Add appendix to table of contents
\newpage
\clearpage

\appendix
\section{Proof of Eq.\ref{eq:backward}}
\label{proof}

From Eq.\ref{eq:fixed_point}, we can easily derive:

\begin{align}
\frac{\partial \cL}{\partial \vX_\tin} = \alpha \frac{\partial \cL}{\partial \vX_*}
\Big ( \vI - (1-\alpha) \tA \Big)^{-1}
\end{align}

\noindent According to the chain rule, we have:
\begin{equation}
\frac{\partial \cL}{\partial \vX_\tin} =
 \frac{\partial \cL}{\partial \vX_*}
\frac{d{X_{*}}}{d{X_{\text{in}}}}
\end{equation}
 
\noindent Then, we have:

\begin{equation}
\frac{d{X_{*}}}{d{X_{\text{in}}}}=
\alpha \Big ( \vI - (1-\alpha) \tA \Big)^{-1}
\end{equation}

\begin{equation}
\left(\frac{d{X_{*}}}{d{X_{\text{in}}}}\right)^T  \frac{\partial \cL}{\partial \vX_*} =
\alpha \Big ( \vI - (1-\alpha) \tA \Big)^{-T}
 \frac{\partial \cL}{\partial \vX_*}
\end{equation}

\noindent For simplicity, suppose:
\begin{equation}
y = \frac{\partial \cL}{\partial \vX_*}
\end{equation}

\begin{equation}
G = \alpha \Big ( \vI - (1-\alpha) \tA \Big)^{-T}
 \frac{\partial \cL}{\partial \vX_*}
\end{equation}

\noindent Then:

\begin{equation}
\Big ( \vI - (1-\alpha) \tA \Big)^{T}G=\alpha y
\end{equation}

\begin{equation}
G = (1-\alpha) \tA G + \alpha y
\end{equation}

\noindent Note that this is also a fixed-point equation, which we choose to solve using an iterative procedure similar to forward propagation. This approach yields Eq.\ref{eq:backward}, where $G_{L} = \frac{\partial \cL}{\partial \vX_L} ~~ \approx \frac{\partial \cL}{\partial \vX_*} $.

\begin{align} 
\vG_{l} &= (1-\alpha) \tA \vG_{l+1} + \alpha \frac{\partial \cL}{\partial \vX_L^k},
~\forall l=L-1, \dots, 0
\end{align}

\section{GPT-2 Performance and Efficiency Analysis}
\label{GPT}

In this section, we extend our investigation of LEADING by incorporating decoder-only LMs, which are usually larger than encoder-only models. Specifically, we perform fine-tuning on Cora and ogbn-arxiv datasets using GPT-2. Unlike the encoder-only models, which utilize the CLS (classification) embedding from the last hidden states of fine-tuned language models (LMs) as text embeddings, we adopt a different strategy for decoder-only models. In this case, we utilize the information from the last token, as it encapsulates all the necessary details for prediction, aligning with the generative nature of decoder-only models. Correspondingly, we pad the sequence on the left.

The results shown in Table \ref{tab:gpt-acc} and \ref{tab:gpt-memory} indicate that the proposed LEADING algorithm effectively fine-tunes GPT-2 to achieve better performance, which is consistent with our experiments on other language models in Table~\ref{tab:accuracy-llm}. Regarding the computation cost, LEADING is capable of maintaining computational costs nearly identical to supervised fine-tuning of GPT-2 without graphs. The additional running time arises due to the sequential execution of two pipelines in LEADING, yet this can be effectively mitigated through parallel computing. It incurs significantly less computation overhead or memory cost compared to baselines such as GLEM. We similarly report N/A for GIANT due to the identical rationale analyzed in Section \ref{sec:efficiency}. It's crucial to emphasize that enhancing model size may not be essential for achieving superior performance. The effectiveness of fine-tuning is influenced by a range of factors beyond mere model size.

\begin{table}[ht!]
\caption{LEADING performance with GPT-2}
\vspace{-0.2in}
\label{tab:gpt-acc}
\begin{center}
\renewcommand{\arraystretch}{1.2}
\resizebox{1\linewidth}{!}{%
\begin{tabular}{lcccccc}
&\multicolumn{2}{c}{\bf Cora} &\multicolumn{2}{c}{\bf Arxiv}\\
\toprule
{\textbf{Method}} & \textsc{GCN} & \textsc{GAT} & \textsc{GCN} & \textsc{Rev-GAT} \\
\midrule
Pre-trained GPT-2 & $51.9 \pm 1.8$ & $54.7 \pm 1.3$ & $64.8$ & $66.9$ \\
Supervised-FT GPT-2 & $70.8 \pm 1.8$ & $71.7 \pm 1.9$ & $73.2$ & $73.8$ \\
GLEM(GPT-2) & $70.8 \pm 1.8$ & $71.7 \pm 1.9$ & $74.0$ & $75.1$ \\
LEADING(GPT-2) & $\mathbf{80.5 \pm 2.3}$ & $\mathbf{81.5 \pm 1.8}$ & $\mathbf{74.7}$ & $\mathbf{75.9}$ \\
\bottomrule
\end{tabular}
}
\end{center}
\end{table}

\begin{table}[!htb]
\caption{Scalability comparison with GPT-2 on ogbn-arxiv}
\vspace{-0.3in}
\label{tab:gpt-memory}
\begin{center}
\renewcommand{\arraystretch}{1.2}
\setlength{\arrayrulewidth}{1.5pt}
\resizebox{0.45\textwidth}{!}{%
\begin{tabular}{lcc}
\\ \toprule
\textbf{\textsc{Methods}} & \multicolumn{1}{c}{\bf Memory(GB)} & \multicolumn{1}{c}{\bf Running Time(S)} \\
\midrule
Supervised-FT GPT-2       & 26.8 & 15555\\
\midrule
GIANT (GPT-2)    & N/A & N/A  \\
GLEM (GPT-2)     & 26.8 &  82930\\
\midrule
LEADING (GPT-2)    & 27.1 & 27920\\
\bottomrule
\end{tabular}
}
\end{center}
\end{table}

\section{Link Prediction}
\label{link}

To emphasize the effectiveness of our LEADING algorithm, we conducted additional experiments on link prediction tasks. We primarily adhere to the implementation in PyTorch Geometric \footnote{\url{https://github.com/pyg-team/pytorch_geometric/tree/master/examples}} and run the proposed algorithm on two widely used datasets: the small dataset Cora and the large dataset OGBL-Citation2. For Cora, We partition the links into distinct sets for training, validation, and testing. This division is performed with two distinct split ratios: (1) a low ratio of 10/30/60 for train/valid/test, and (2) a high ratio of 85/5/10 for the same sets. We use GCN as the downstream GNN. For OGBL-Citation2, since it was customized in existing literature\cite{duan2023simteg}, we adhered to the same experiment settings and utilized a two-layer GraphSAGE model for the downstream GNN.
To prioritize faster execution and simplicity, we choose DistillBert and another popular lightweight deep text embedding models Sentence-BERT as our language models. As indicated in Table \ref{tab:link} and \ref{citation2}, LEADING exhibits performance advantages over the baselines, especially in scenarios with limited training data. These results reinforce our paper's observations regarding node classification and underscore the remarkable versatility of our proposed method.

\begin{table}[ht!]
\caption{LEADING performance of link prediction on Cora}
\label{tab:link}
\begin{center}
\renewcommand{\arraystretch}{1}
\vspace{-0.15in}
% \begin{small}
% \begin{sc}
\resizebox{0.8\linewidth}{!}{%
\begin{tabular}{lcccccc}
\toprule
{\textbf{Method}} & \textsc{Low} & \textsc{High} \\
\midrule
Shallow Embedding  & 79.7 & 94.9 \\
\midrule
Pre-trained DistillBERT  & 64.7 &  68.7  \\
Pre-trained SBERT  & 80.3 & 95.0   \\
\midrule
Supervised-FT DistillBERT & 66.3 &  89.4  \\
Supervised-FT SBERT & 80.8 & 95.3   \\
\midrule
LEADING(DistillBERT) & 81.8  &  95.2  \\
LEADING(SBERT) &  \textbf{83.3} & \textbf{95.7}   \\
\bottomrule
\end{tabular}
}
\end{center}
% \end{table*}
\end{table}

\begin{table}[!htb]
\centering
\renewcommand{\arraystretch}{1.2}
\caption{LEADING performance of link prediction on OGBL-Citation2}
\label{citation2}
\vspace{-0.15in}
\resizebox{1\linewidth}{!}{%
\begin{tabular}{l|cccc}
\toprule
\multirow{1}{*}{\textbf{Method}} & \multirow{1}{*}{\textbf{Shallow}} & \multirow{1}{*}{\textbf{Pre-trained}} & \multirow{1}{*}{\textbf{Supervised-FT}} & \multirow{1}{*}{\textbf{LEADING}} \\
\midrule
SBERT& 77.3 & 81.8 & 83.1 & \textbf{85.2} \\
\bottomrule
\end{tabular}
}  
\end{table}

\section{PEFT}
\label{peft}

Additionally, we illustrate the versatility of our proposed LEADING approach by successfully integrating it with existing Parameter Efficient Fine Tuning (PEFT) approaches. Specifically, we opt for LoRA, a proven and effective method for fine-tuning the model to enhance performance. We choose to tune query, key, value, dense layer and linear layer using LoRA. From the Table \ref{tab:lora}, we observe that LoRA can further enhance performance in conjunction with our proposed LEADING algorithm.

\begin{table}[!htb]
\caption{Prediction accuracy (\%) with PEFT.}
\vspace{-0.15in}
\label{tab:lora}
\begin{center}
\renewcommand{\arraystretch}{1}
\setlength{\arrayrulewidth}{1pt}
\resizebox{0.5\textwidth}{!}{%
\begin{tabular}{lcccc}
& \multicolumn{2}{c}{\bf Cora}  &\multicolumn{2}{c}{\bf Pubmed}
\\ \toprule
\textbf{\textsc{Methods}} & \multicolumn{1}{c}{\bf GCN} & \multicolumn{1}{c}{\bf GAT} &\multicolumn{1}{c}{\bf GCN} & \multicolumn{1}{c}{\bf GAT}\\
\midrule
% Default Feature
Shallow Embedding  & $82.0 \pm 0.7$ & $82.3 \pm 0.7$ & $78.9 \pm 1.0 $ & $77.7 \pm 0.9 $\\
\midrule
% FT BERT Feature      
Supervised-FT BERT  & $77.3 \pm 1.7$ & $78.2 \pm 1.4$ & $68.6 \pm 1.8$ & $68.6 \pm 1.4$ \\
% FT DeBERTa Feature 
\midrule
LEADING (BERT)   & $80.5 \pm 0.4$ & $81.6 \pm 0.3$ & $79.1 \pm 0.5$ & $79.0 \pm 1.0$ \\
LEADING (LoRA)   & $\mathbf{82.3 \pm 0.5}$ & $\mathbf{82.5 \pm 0.6}$ & $\mathbf{81.0 \pm 0.6}$ & $\mathbf{79.5 \pm 0.8}$\\
\bottomrule
\end{tabular}
}
\end{center}
\end{table}

\section{Statistical Significance Tests}
\label{significant}
Given the close performance on ogbn-arxiv, to clearly demonstrate the substantial performance enhancement, we conducted five sets of experiments on both our proposed algorithm and the top-performing baseline, GLEM, and then performed statistical significance tests. The p-value is 8.38e-06 for Rev-GAT and 1.64e-05 for GCN, much less than the commonly used desired significance level 0.01 and highly significance level 0.001. It clearly show that we can reject the null hypothesis. The difference is highly significant.

\begin{table}[!htb]
\caption{Statistical significance test on ogbn-arxiv}
\vspace{-0.15in}
\label{tab:significant}
\begin{center}
\renewcommand{\arraystretch}{1.2}
\setlength{\arrayrulewidth}{1.5pt}
\resizebox{0.4\textwidth}{!}{%
\begin{tabular}{lcccc}
& \multicolumn{2}{c}{\bf GCN}  &\multicolumn{2}{c}{\bf REV-GAT} 
\\ \toprule
\textbf{\textsc{Runs}} & \multicolumn{1}{c}{\bf GLEM} & \multicolumn{1}{c}{\bf LEADING} &\multicolumn{1}{c}{\bf GLEM} & \multicolumn{1}{c}{\bf LEADING} \\
\midrule
% Default Feature
1  &75.90  &76.18 &76.90 &77.42   \\
2  &75.95  &76.10 &76.80 &77.61   \\
3  &75.93  &76.08 &77.00 &77.52   \\
4  &75.86  &76.13 &76.85 &77.34  \\
5  &75.87  &76.12 &76.78 &77.46   \\
\midrule
p-value & \multicolumn{2}{c}{1.64e-05}  &\multicolumn{2}{c}{8.38e-06} \\
\bottomrule
\end{tabular}
}
\end{center}
\end{table}

\section{High Labeling Rate}
\label{high}

Additionally, we include an extra experiment to demonstrate that the proposed approach is also effective in scenarios with a high labeling rate for the Cora dataset. To verify this, we set 60\% of the nodes as training set, 20\% as the validation set, and the remaining 20\% as the test set. The DeBERTa trained by LEADING clearly outperforms all the baselines, which verifies our conclusions as shown in Table \ref{tab:high rate}.

\begin{table}[!htb]
\caption{Prediction accuracy (\%) on Cora in high labeling rate setting.}
\vspace{-0.15in}
\label{tab:high rate}
\begin{center}
\renewcommand{\arraystretch}{1}
\setlength{\arrayrulewidth}{0.8pt}
\resizebox{0.4\textwidth}{!}{%
\begin{tabular}{ccc}
\toprule
\textbf{\textsc{Methods}} & \multicolumn{1}{c}{\bf GCN} & \multicolumn{1}{c}{\bf GAT} \\
\midrule
% Default Feature
Shallow Embedding  & $90.9 \pm 2.7$ & $90.6 \pm 3.0$ \\

Pre-trained DeBERTa  & $65.9 \pm 2.0 $ & $79.7 \pm 3.2$\\

Supervised-FT DeBERTa & $85.9 \pm 2.3$ & $86.5 \pm 1.9$\\

GLEM (DeBERTa)  & $89.1 \pm 0.7$ & $89.0 \pm 0.6$\\

LEADING (DeBERTa)   & $ \mathbf{92.5 \pm 2.3}$ & $ \mathbf{93.2 \pm 1.8}$\\
\bottomrule
\end{tabular}
}
\end{center}
\end{table}

Compared with Table \ref{tab:accuracy-llm}, we point out that the performance of GLEM highly depends on the labeling ratio. In all of our experiments (for all LM backbones such as
DeBERTa and GPT-2), GLEM works pretty well for the cases of high labeling ratios. However, The underlying reason for the above phenomenon is that GLEM adopts a two-stage approach (instead of the end-to-end training as our LEADING algorithm): (1) generate pseudo labels and (2) supervised fine-tuning of LMs on the generated pseudo labels. Therefore, the effectiveness of supervised tuning of LMs on the generated pseudo labels heavily relies on the quality of those pseudo labels. In the common low labeling data split of Cora, the quality of pseudo labels is low such that fine-tuning of GLEM using these low-quality pseudo labels will even harm the accuracy.

\section{Comparison with other joint training paradigms}
\label{e2e}

Recently, several joint training paradigms have been proposed. We further include one classical model, GraphFormers\cite{yang2021graphformers}, and one state-of-the-art model, Grenade\cite{li2023grenade}, in our performance comparison. Specially, GraphFormers facilitate the co-training of pre-trained LMs and GNNs by injecting GNN layers into the language model's architecture. GRENADE jointly optimizes GNN and LM encoders by employing graph-centric contrastive learning and dual-level graph-centric knowledge alignment. We follow the same experimental settings as in their original papers. Note that, since GraphFormers are originally designed for link prediction, we keep our implementation settings for node classification the same as in \cite{zhao2022learning}. The results are shown in Table \ref{tab:e2e}. 

We can observe that the proposed LEADING algorithm outperforms these two joint training paradigms on both datasets. We conclude that this is because their designs are not as effective as our LEADING algorithm. These results demonstrate the superiority of our design.

\begin{table}[!h]
\caption{Prediction accuracy (\%) comparison with other joint training paradigms}
\vspace{-0.15in}
\label{tab:e2e}
\begin{center}
\renewcommand{\arraystretch}{1}
\setlength{\arrayrulewidth}{0.8pt}
\resizebox{0.4\textwidth}{!}{%
\begin{tabular}{ccc}
\toprule
\textbf{\textsc{Methods}} & \multicolumn{1}{c}{\bf Arxiv} & \multicolumn{1}{c}{\bf Products} \\
\midrule
% Default Feature
GraphFormers  & $72.8 \pm 0.20$ & $74.7 \pm 0.16$ \\

Grenade  & $75.0 \pm 0.19 $ & $83.1 \pm 0.56$\\

LEADING   & $ \mathbf{76.0 \pm 0.05}$ & $ \mathbf{84.1 \pm 0.32}$\\
\bottomrule
\end{tabular}
}
\end{center}
\end{table}

\section{Hyperparameters}
\label{hyperparameters}
In this section, we investigate the impact of one important hyperparameter in LEADING: varying the batch size of the second pipeline on the final performance. This parameter directly influences the encoding times and the frequency at which the embeddings in memory are refreshed, as introduced in Section \ref{sec:decouple}. We conducted experiments on ogbn-arxiv, using a 2-layer APPNP as the GNN backbone and DeBERTa as the language model. We keep the batch size of the first pipeline constant at 128 and vary the batch size of the second pipeline. The results are shown in the following table: 

\begin{table}[!h]
\caption{Prediction accuracy (\%) comparison with varying batch sizes}
\vspace{-0.15in}
\label{tab:bs}
\begin{center}
\renewcommand{\arraystretch}{1}
\setlength{\arrayrulewidth}{0.8pt}
\resizebox{0.5\textwidth}{!}{%
\begin{tabular}{ccc}
\toprule
\textbf{\textsc{Batch Size}} & \multicolumn{1}{c}{\bf Accuracy} & \multicolumn{1}{c}{\bf Encoding Times per Node} \\
\midrule
% Default Feature
128  & 73.6 & 2 \\

256  & 73.9 & 3\\

512   & 73.9 & 5\\
\bottomrule
\end{tabular}
}
\end{center}
\end{table}

The presented results indicate that using a larger batch size in the second pipeline can enhance performance by refreshing the memory more frequently, as it accelerates memory table updates and reduces staleness. However, the improvement is marginal, and a larger batch size incurs additional memory costs. Consequently, we choose to use the same batch size for both the first and second pipelines to maintain good performance while preserving the memory efficiency of our proposed algorithm.

\end{document}